\pdfoutput=1

\documentclass[11pt]{article}

\usepackage{acl}

\usepackage{times}
\usepackage{latexsym}

\usepackage{multirow}
\usepackage{enumitem}

\usepackage{graphicx}
\graphicspath{ {./images/} }

\usepackage[T1]{fontenc}

\usepackage[utf8]{inputenc}

\usepackage{microtype}

\usepackage{xurl}

\title{Collecting high-quality adversarial data for machine reading\\comprehension tasks with humans and models in the loop}

\author{Damian Y. Romero             Diaz,\textsuperscript{\includegraphics[scale=0.16]{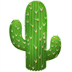},\includegraphics[scale=0.16]{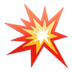}}
        Magdalena Anioł,\textsuperscript{\includegraphics[scale=0.16]{explosion.png}}
        John Culnan\textsuperscript{\includegraphics[scale=0.16]{uofa.png}} \\
        \texttt{damian@explosion.ai\thanks{ \phantom{ } Corresponding author.}, magda@explosion.ai, jmculnan@arizona.edu} \\
        \textsuperscript{\includegraphics[scale=0.16]{explosion.png}} Explosion, 
        \textsuperscript{\includegraphics[scale=0.16]{uofa.png}} University of Arizona}

\begin{document}
\maketitle
\begin{abstract}

We present our experience as annotators in the creation of high-quality, adversarial machine-reading-comprehension data for extractive QA for Task 1 of the First Workshop on Dynamic Adversarial Data Collection (DADC). DADC is an emergent data collection paradigm with both models and humans in the loop. We set up a quasi-experimental annotation design and perform quantitative analyses across groups with different numbers of annotators focusing on successful adversarial attacks, cost analysis, and annotator confidence correlation. We further perform a qualitative analysis of our perceived difficulty of the task given the different topics of the passages in our dataset and conclude with recommendations and suggestions  that might be of value to people working on future DADC tasks and related annotation interfaces.

\end{abstract}

\section{Introduction}

We present quantitative and qualitative analyses of our experience as annotators in the machine reading comprehension shared task for the First Workshop on Dynamic Adversarial Data Collection.\footnote{\url{https://www.aclweb.org/portal/content/call-participation-shared-task-first-workshop-dynamic-adversarial-data-collection-dadc}}. The shared task was a collection of three sub-tasks focused on the selection of excerpts from unstructured texts that best answer a given question (extractive question-answering). The sub-tasks included: (A) the manual creation of question-answer pairs by human annotators, (B) the submission of novel training data (10,000 training examples), and (C) the creation of better extractive question-answering models. In this paper, we focus on our participation in the the manual creation of question-answer pairs task dubbed as "Track 1: Better Annotators".


\begin{table*}
    \centering
    \begin{tabular}{lcccc}
    \hline
    \multirow{2}{*}{\bf Annotation Result } & \multirow{2}{*}{\bf Total} & {\bf Single Annotator} & {\bf Two-Annotator} & {\bf Three-Annotator} \\
    & & {\bf Sessions} & {\bf Sessions} & {\bf Sessions} \\
    \hline 
    Model fooled & 45 & 21 & 19 & 5 \\
    Model not fooled & 43 & 22 & 13 & 8 \\
    False negative & 10 & 5 & 4 & 1 \\
    False positive & 2 & 1 & 1 & 0 \\
    \hline
    Total & 100 & 49 & 37 & 14 \\
    \hline 
    \end{tabular}
    \caption{Overall annotation results before verification.}
    \label{tab:overall-results}
\end{table*}

Machine reading comprehension (MRC) is a type of natural language processing task that relies in the understanding of natural language and knowledge about the world to answer questions about a given text \citep{rajpurkar-etal-2016-squad}. In some cases, state-of-the-art MRC systems are close to or have already started outperforming \emph{standard} human benchmarks
\citep{dzendzik-etal-2021-english}. However, models trained on \emph{standard} datasets (i.e., collected in non-adversarial conditions) do not perform as well when evaluated on adversarially-chosen inputs \citep{jia-liang-2017-adversarial}.

To further challenge models and make them robust against adversarial attacks, researchers have started creating adversarial datasets which continuously change models as they grow stronger. Dynamic Adversarial Data Collection (DADC) is an emergent data collection paradigm explicitly created for the collection of such adversarial datasets. In DADC, human annotators interact with an adversary model or ensemble of models in real-time during the annotation process \citep{bartolo2020beat} to create examples that elicit incorrect predictions from the model \citep{kaushik2021efficacy}. DADC allows for the creation of increasingly more challenging data as well as improved models and benchmarks for adversarial attacks \citep{dua2019drop,kaushik2021efficacy,nie2019adversarial}.

There is evidence that data collected through adversarial means is distributionally different from standard data. From a lexical point of view, \citet{kaushik2021efficacy} note that “what-” and “how-” questions dominate in adversarial data collection (ADC) as opposed to “who-” and “when-” questions in the standard datasets. In the context of reading comprehension, DADC has been championed by \citet{bartolo2020beat}, who observe that DADC QA datasets are generally syntactically and lexically more diverse, contain more paraphrases and comparisons, and often require multi-hop inference, especially implicit inference.

Apart from corpus analyses, researchers have also noted certain limitations of the DADC paradigm. For instance, \citet{kiela2021dynabench} note that annotators overfitting on models might lead to cyclical progress and that the dynamically collected data might rely too heavily on the model used, which can potentially be mitigated by mixing in standard data. Similarly, \citet{kaushik2021efficacy} find that DADC models do not respond well to distribution shifts and have problems generalizing to non-DADC tests.

\paragraph{Contributions}

In this paper, we present our experience as annotators in the reading comprehension shared task for the First Workshop on Dynamic Adversarial Data Collection. Through quantitative and qualitative analyses of a quasi-experimental annotation design, we discuss issues such as cost analysis, annotator confidence, perceived difficulty of the task in relation to the topics of the passages in our dataset, and the issues we encountered while interacting with the system, specifically in relationship with the commonly-used F1 word-overlap metric. We conclude with recommendations and suggestions  that might be of value to people working on future DADC tasks and related annotation interfaces.

\section{Task Description}

Track 1 of the First Workshop on Dynamic Adversarial Data Collection consisted in generating 100 reading comprehension questions from a novel set of annotation passages while competing against the current state-of-the-art QA model \citep{bartolo2021improving}, which would remain static throughout the task. Through  Dynabench \citep{kiela2021dynabench},\footnote{\url{https://dynabench.org/}} an annotation platform specialized in DADC, annotators would create model-fooling questions that could be answered with a continuous span of text. Successful attacks required the annotators to provide explanations of the question and a hypothesis for the model's failure. These were then subject to a \emph{post hoc} human validation.

\subsection{F1 metric and false negatives}

During our participation, we discovered two issues with the implementation of the metric used in Dynabench to decide whether the model had been fooled or not.

Dynabench uses a word-overlap metric to calculate the success of the model(s)’ responses against those selected by the annotators \citep{kiela2021dynabench}. This metric is calculated as the F1 score of the overlapping words between the answer selected by the annotators and the answer predicted by the model, where model responses with a score above 40\% are labeled as a successful answer for the model. For example, the answer “New York” would be considered equivalent to the answer “New York City" \citep{bartolo2020beat}.

In practice, we observed that the F1 metric led to many false negatives,\footnote{Notice that, from a model-evaluation perspective, these would be considered false positives.} or, in other words, to answers that were considered unsuccessful attacks from the annotators when, in reality, the model was wrong. This happened in two different circumstances. First, in the form of incomplete answers where critical information was missing from the model's answer, and the answer was still considered equivalent due to a sufficient word overlap, as in example A from Table~\ref{tab:qa-examples}.

\begin{table*}[h!]
    \centering
    \columnsep 1cm
    \begin{tabular}{lll}
    \hline
    {\bf Question} & {\bf Model's answer} & {\bf Annotators' answer} \\
    \hline 
     \multirow{2}{*}{A) What was Eric Fellner working on?} & \multirow{2}{*}{Tinker Tailor Soldier Spy} & \emph{A sequel to Tinker Tailor} \\
    & & Soldier Spy \\
    \\
    B) At what times is the eastern & \multirow{2}{*}{5:00 am to 3:30 pm} & 5:00 am to \emph{6:00 pm,} \\
    walkway open to pedestrians only? & & \emph{or 9:00 pm during DST} \\
    \hline 
    \end{tabular}
    \caption{Examples of questions, model answers, and annotators' answers in the data creation procedure. All question-answer examples are adapted from the Dynabench dataset.}
    \label{tab:qa-examples}
\end{table*}


In this case, since "Tinker Tailor Soldier Spy" is a movie, it cannot be said that the first movie and the sequel are equivalent. This behavior was so common that we decided to turn it into an adversarial-attack strategy by forcing the model to provide full answers, which it could not do because of its strong bias towards short answers. For example, we asked questions such as "What is the \emph{full location} of the plot of the TV show?", for which the model tended to answer with the bare minimum of information due to being trained using the F1 word-overlap metric.

In other cases, the model selected a different text span than the one selected by the annotators, as in example B from Table~\ref{tab:qa-examples}.

In this case, not only is the model’s answer incomplete but \emph{3:30 pm} and \emph{6:00 pm} have entirely different meanings. Cases such as the one above occurred in passages that had two very similar strings in the text. In these cases, the F1 metric lead Dynabench to score in favor of the model even when the answer was incorrect. We believe that the answer provided by the annotators, in cases where annotators are hired as experts in a given domain, should be considered a gold standard subject to the validation process. In other cases, when annotations come from crowdsourcing platforms, the F1 metric could be more adequate.

\section{Methodology}

Our annotator roster consisted of three annotators with postgraduate degrees in linguistics and natural language processing. One of the annotators spoke English as a first language, while the other two were proficient speakers of English as a second language who completed their graduate degrees in English-speaking universities. For the annotation process, we set up a quasi-experimental design using convenience sampling where approximately half of the annotations would be performed by a single annotator (n=49), and the other half would be performed synchronously by a group of two or more annotators (n=51). Because the annotators live in different time zones, annotator groups did not remain consistent across group sessions.

During the annotation task, the platform randomly picked a passage, usually of the length of a short paragraph (of about 160 words on average) from different topics. Annotators could then choose to create questions for that passage or skip it entirely. Annotators skipped passages when we agreed that it would be difficult to create even a single question to fool the model.\footnote{We did not keep track of the passages we skipped.} Table~\ref{tab:overall-results} contains our overall annotation results by the number of annotators. We report our results using the following typology:

\bigskip
\noindent
{\bf Model fooled:} Items marked by Dynabench as successful annotations.

\noindent
{\bf Model not fooled:} Items marked by Dynabench as unsuccessful annotations.

\noindent
{\bf False negatives:} Instances where the model was fooled, but Dynabench marked them as not fooled.\footnote{Mainly due to F1 score problems.}

\noindent
{\bf False positives:} Items marked by Dynabench as successful annotations but deemed unsuccessful by the annotators.\footnote{There are two of these in the dataset and they were products of mistakes the annotators made when selecting the answers on Dynabench that lead to a mismatch between the question asked and the answer given to the model.}

\bigskip
Even though the limited number of examples does not allow us to draw any strong conclusions about the annotation task, we find our analyses worth presenting as a preliminary step for other annotators to further reflect on the annotation process during the planning stages of any DADC task.\footnote{The code for our analyses can be found at \url{https://github.com/fireworks-ai/conference-papers/tree/master/naacl-dadc-2022}}

\subsection{Model fooled ratio by annotator group}

\begin{figure*}
    \includegraphics[width=16cm]{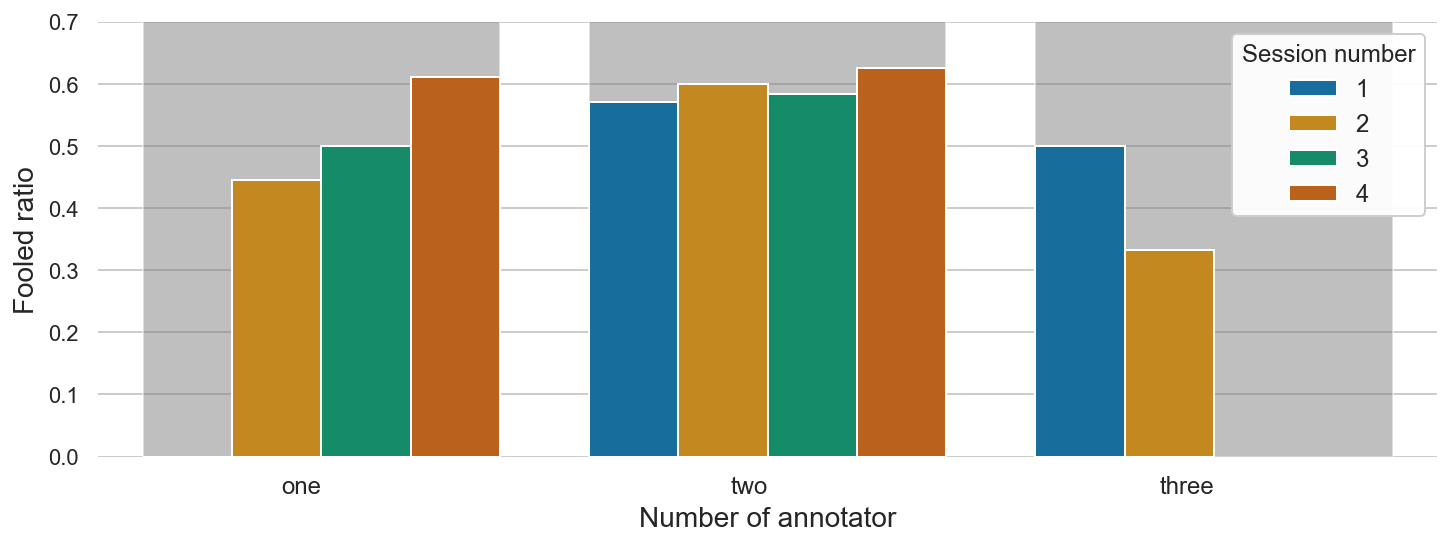}
    \centering
    \caption{Model fooled ratio by annotator group by session. False negatives and false positives are excluded. Missing sessions had a score of zero.}
    \label{fig:model_fooled_ratio}
\end{figure*}

In order to capture if we as annotators are increasingly improving our model-fooling skills, we investigate the progression of the “model fooled / model not fooled” ratio throughout the annotation sessions. Figure~\ref{fig:model_fooled_ratio} summarizes the results. 

For the single-annotator group, the progression seems apparent with a progressive fool ratio of 0, .44, .50, and .61. Sessions with two annotators do not have a clear progression (0.57, 0.60, 0.58, and 0.62), which may be because annotators did not remain the same in each session. The worst performance happened with the three-annotator sessions (0.50 and 0.33), which indicates a possible high degree of disagreement across annotators.

\begin{figure}
    \includegraphics[width=7.5cm]{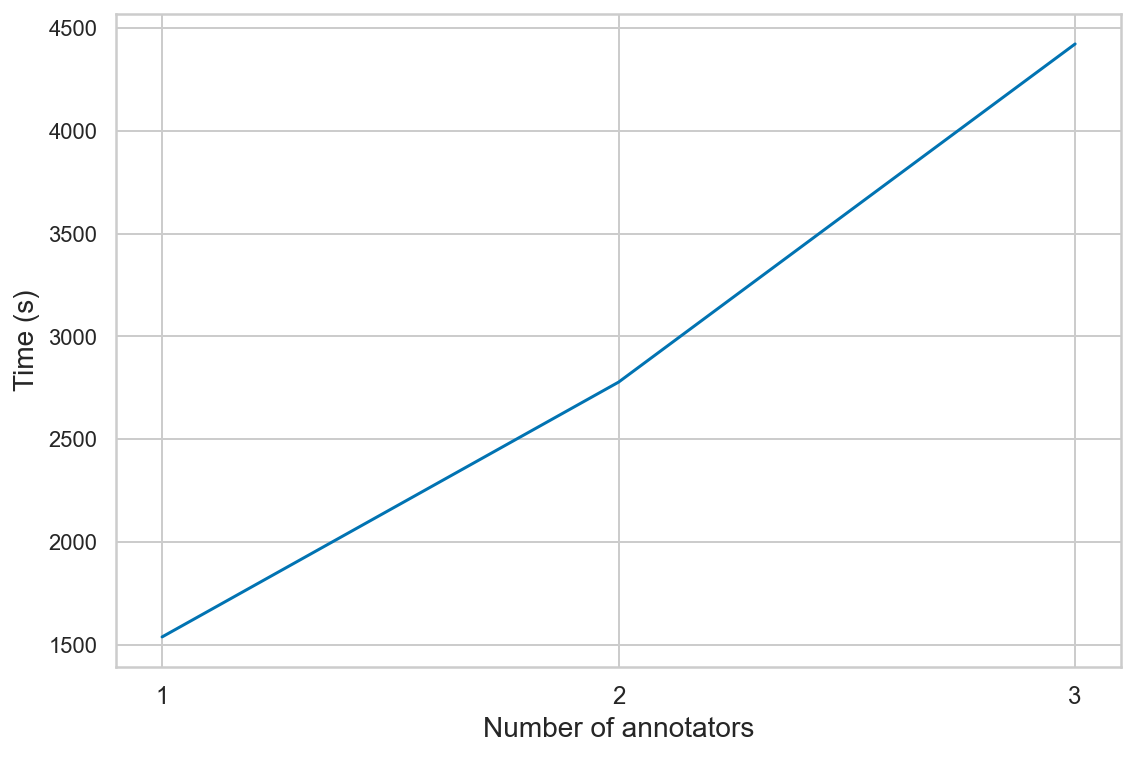}
    \centering
    \setlength{\belowcaptionskip}{-15pt}
    \caption{Mean time in seconds spent per annotator for every successful adversarial attack across groups with different annotators.}
    \label{fig:time_per_fooled}
\end{figure}

\subsection{Annotation costs}

We investigate the efficiency of the different annotator groups by calculating the mean time per successful adversarial attack. Formally, we define annotation group efficiency \(E(g)\) as:

\[E(g) = {\sum_n^k t_n \times a_n \over N} \]

Where \(t_n\) is the total time in seconds spent in annotation session \(n\), \(k\) is the total number of sessions for annotator group \(g\), \(a_n\) is the number of annotators in the session, and \(N\) is the total number of successful adversarial attacks across all the annotation sessions for group \(g\). Table~\ref{tab:annotation-costs} shows annotation efficiency in seconds.

\begin{table}[h!]
    \renewcommand{\arraystretch}{1.2} 
    \centering
    \begin{tabular}{cc}
    \hline
    {\bf Number of } & {\bf Mean time (s) per } \\
    {\bf annotators in group } & {\bf successful example } \\
    \hline 
    {\bf 1} & {\bf 1537.04} \\
    2 & 2777.58 \\
    3 & 4423.80 \\
    \hline 
    Total mean time & 8738.42 \\
    \hline 
    \end{tabular}
    \caption{Mean time in seconds spent per annotator for every successful adversarial attack across groups with different annotators.}
    \label{tab:annotation-costs}
\end{table}

The single-annotator group took 8h 58' 58'' to create 21 model-fooling examples, rendering efficiency of 25' 37'' per successful attack. For the group annotations, the two-annotator sessions took 14h 39' 34'' to create 19 model-fooling examples, with an efficiency of 46' 17'', while the three-annotator sessions took 6h 8' 39'' to create five successful examples with an efficiency of 1h 13'. The total time spent on the task was 29h 46'' 11'. Figure~\ref{fig:time_per_fooled} shows (in seconds) how the time increment is almost linear.

\subsection{Confidence scores}

Lastly, to better understand why annotation times took longer when working in groups, we investigate the level of confidence agreement between annotators via correlation. To measure confidence agreement, annotators individually logged in confidence scores for all of the 100 questions in our dataset. The scores range between 0 and 3 points, with three being entirely confident that they would fool the model.

\begin{figure}[h!]
    \includegraphics[width=7.5cm]{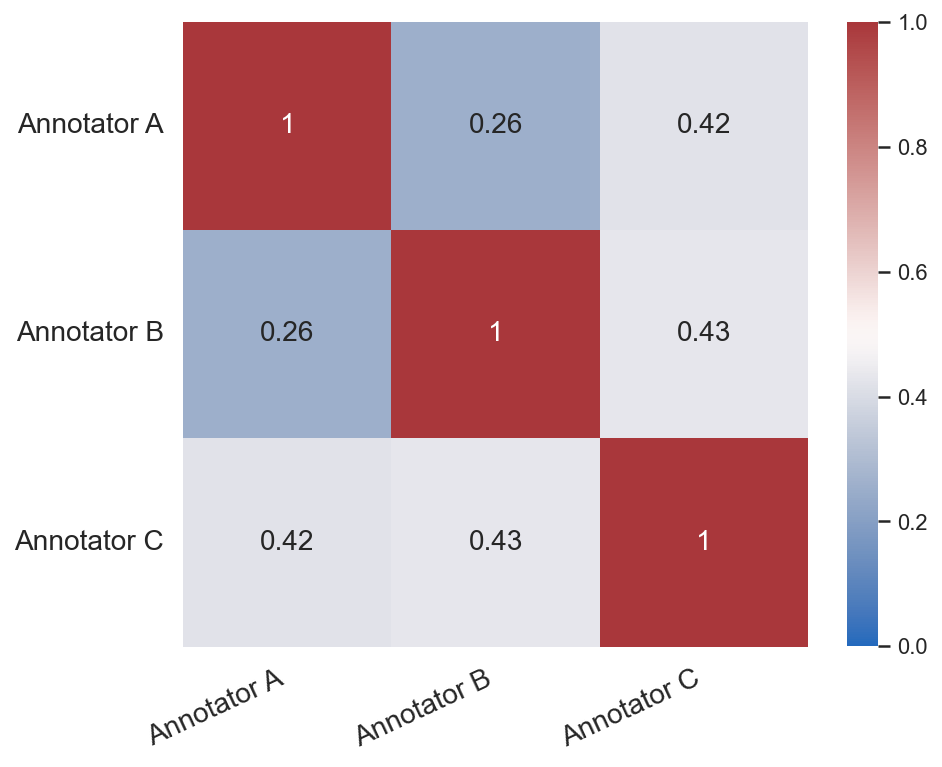}
    \centering
    \caption{Correlation heatmap of annotators’ confidence metrics through the full dataset.}
    \label{fig:confidence_heatmap}
\end{figure}

We first test our data for normality using the "normaltest" function of the Python SciPy library \citep{virtanen2020scipy}. After ensuring that normality tests came out negative across all annotators' ratings (p < 0.001), we used the Spearman rank correlation test (Figure ~\ref{fig:confidence_heatmap}) as implemented in the Python Pandas library \citep{mckinney2010data, reback-pandas}.

The fact that correlation coefficients range from weak to moderate supports our view that the lower efficiency in annotation costs might be due to differences in how annotators perceive how the model will evaluate their questions. This could lead to more debate during the synchronous annotation sessions. The lack of exponential time increase when more annotators are present, as was the case of the sessions with three annotators, may be due to the fact that annotators were often tired of the feeling of low-productivity of the sessions and were, at times, willing to risk questions without fully debating them.

\section{Qualitative Analysis}


The relative difficulty of a dynamic adversarial dataset creation task may vary partly as a function of the genre and specific topic of the text passages from which question-answer pairs are drawn. During the shared task, Dynabench randomly assigned passages for the creation of question-answer pairs, revealing several important aspects of this challenge. Topics of the passages used in our data vary as shown by the success by topic scores in  Table~\ref{tab:model-fooled-per-category}.

\begin{table*}[]
    \small
    \centering
    \renewcommand{\arraystretch}{1.5} 
    \begin{tabular}{lccccccccc}
    \hline 
    {\bf } & \multirow{2}{*}{\bf{Comics}} & \multirow{2}{*}{\bf{History}} & \multirow{2}{*}{\bf{Literature}} & \multirow{2}{*}{\bf{Music}} & \multirow{2}{*}{\bf{Science}} & \multirow{2}{*}{\bf{Sports}} & \bf{TV and} & \multirow{2}{*}{\bf{Technology}} & \bf{Video} \\
    & & & & & & & \bf{Movies} & & \bf{Games} \\
    \hline 
    Model Fooled & 1 & 6 & 5 & 5 & 4 & 8 & 9 & 2 & 5 \\
    Total Items & 3 & 16 & 7 & 9 & 4 & 23 & 20 & 4 & 14 \\
    \hline 
    Ratio Fooled & 0.33 & 0.37 & \textbf{0.71} & \textbf{0.55} & \textbf{1.00} & 0.35 & 0.45 & 0.50 & 0.36 \\
    \hline
    \end{tabular}
    \caption{Number of times our questions fooled the model out of the total number of questions we generated for each passage topic in our dataset. False negatives and false positives are included in the total number of items.}
    \label{tab:model-fooled-per-category} 
\end{table*}

Our more successful questions came from music, science, and technology topics. On the one hand, we are more familiar with these topics than comics, sports, and video games. Furthermore, the paragraphs in literature and music tended to be more narrative in nature which, we believe, also made it easier for us to process them and create better questions. Data-heavy, enumeration-based paragraphs typical of sports, history, and TV and movies topics proved more challenging for the creation of model-fooling questions. Still, further examination is necessary to understand each of these possibilities separately.

A closer examination of the DADC task included evaluating the success of different strategies for creating questions. Overall, the model successfully answered questions about dates and names, as well as questions that could be answered with a single short phrase, especially if that phrase was produced as an appositive. For example, asking “Which political associate of Abraham Lincoln was aware of his illness while traveling from Washington DC to Gettysburg?” allowed the model to select a name as the answer, which it did with a high degree of success, even when multiple distractor names appeared in the same paragraph. On the other hand, formulating questions that required longer answers, especially questions that asked for both “what” and “why”, frequently fooled the model. Furthermore, requiring references to multiple non-contiguous portions of the passage to make predictions also often fooled the model. Still, using synonymous words or phrases or similar sentence structures to the critical portions of the passage allowed the model to make correct predictions, even when these other strategies may have fooled it under different circumstances.




\section{Discussion}

Based on the experience with DADC shared task Track 1, we recommend several strategies to improve the efficiency of data collection.

\subsection{Experimenting with the task}

We found that allowing annotators to run "dry" trials before starting data collection, as done by the organizers of the DADC Shared Task, might help them form initial hypotheses about the potential weaknesses of the model and what strategies could be helpful to fool it, e.g., targeting different capabilities such as NER or coreference resolution. Additionally, it could be possible that once annotators are familiarized with the task and understand what examples have a better chance of fooling the model, productivity between multiple annotators might increase as their confidence starts to align.

\subsection{Familiarity with the domain}

We believe it may be significantly easier to come up with good-quality questions if the annotators are familiar with the domain of the contexts. Not only can they read and understand the paragraphs faster, but it is easier to abstract from the immediate context and, thus, ask more challenging questions. Annotation managers of campaigns with heterogeneous datasets might want to consider recruiting experts for technical or specific sub-domains and crowdsourced workers for those texts consisting of general knowledge.

\subsection{Having a list of strategies}

Keeping a rough track of what annotation strategies worked best proved useful to us during annotation. As an example of the types of strategies that annotators can keep track of and  implement, below we list the strategies we favored for creating model-fooling questions.

\begin{enumerate}[itemsep=-2pt]
    \item Play with the pragmatics of the question, for instance: 
    
    \noindent
    {\bf Question:} What is the full location of the plot of this TV show?
    
    \noindent
    {\bf Annotators' answer:} A mysterious island somewhere in the South Pacific Ocean
    
    \noindent
    {\bf Model's answer:} South Pacific Ocean
    
    \noindent
    {\bf Explanation:} The model is biased towards the shortest answer, which does not always cover the information human need as an answer (Grice's principle of quantity)
    
    \item Change the register, e.g., ask a question as a five-year-old would.
    \item Whenever possible, ask a question that requires a holistic understanding of the whole paragraph (not just a particular sentence).
    \item Ask questions that require common sense reasoning, e.g., about the causes and effects of events.
    \item Ask questions about entities that appear multiple times or have multiple instances in the paragraph. 
\end{enumerate}

\subsection{Discussing created prompts with other annotators}

Another practice that can help is to work in teams whereby annotators would come up with questions in isolation and then rank and further modify them in a brainstorming session. In our experience, having two annotators in one session was almost as efficient as having only one annotator and made the task more engaging,ludic and, consequently, less tedious, potentially reducing the risk of burnout syndrome.

\subsection{Suggestions for future DADC annotation interfaces}

Because DADC annotation applies to NLI and QA datasets (Kiela et al., 2021), we believe that specific considerations would be necessary for future projects that make use of a dedicated DADC interface, including the following:

\begin{itemize}
    \item Given that one of the issues we observed was that many of the successful questions were unnatural and, thus, probably, not helpful for real-life scenarios, annotation platforms could include a naturality score to encourage annotators to create data that will be used in real-world scenarios.
    \item Because the word-overlap F1 threshold seems to vary depending on what is enough information and the appropriate information needed to answer specific questions, we believe that a language model could be trained to replace or aid the F1 metric.
    \item Annotation interfaces could also help annotators by displaying relevant visualizations of the training data so that annotators could try to fool the model in those cases where the model contains little or no data. For example, Bartolo et al. (2020, pp. 17-19) provide bar plots and sunburst plots\footnote{The dataset statistics are only available in the pre-print version of their paper, available at: https://arxiv.org/abs/2002.00293} of question types and answer types for each of their modified datasets. We believe that displaying such visualizations to the annotators in a targeted way could potentially increase their performance while also helping balance the creation of datasets.
    \item Finally, we believe that augmenting the interface with functionality for storing and managing annotation strategies such as the ones mentioned above, together with their rate of effectiveness, could make the annotation process more efficient.
\end{itemize}

\subsection{Final considerations}

Beyond any of the suggestions above, we believe that the DADC has certain limitations that annotation campaigns should be aware of.

In our experience in the context of this extractive QA task, we found it extremely difficult to fool the model, primarily because of its powerful lexical and syntactic reasoning capabilities. This was partly because we were constrained to create questions that a continuous string of text could answer. In many cases, we relied on very complex lexical and syntactic inferences (e.g., violating syntactic islands), which often led to unnatural questions that were unlikely to appear in the real world.

The problem of creating model-fooling examples has already been acknowledged in previous research (Bartolo et al., 2020; Kiela et al., 2021) and is generally addressed by either providing question templates to edit or mixing questions from other "more naturally-distributed" datasets. We want to draw the attention of anyone wishing to apply DADC to their problem of this risk.

Kiela et al. (2021) note that applying DADC for generative QA is not a straightforward task. However, it is perhaps in generative tasks where DADC could offer more value. Given how powerful the SOTA models are, the DADC extractive datasets seem doomed to be eventually skewed towards long and unnatural examples. This is one of ours: "Despite knowledge of which fact does Buffy still allow herself to pass at the hands of an enemy, protecting the one to whom the fact relates by doing so?"

\section*{Acknowledgements}

We thank the organizers and sponsors of the first DADC shared task, especially Max Bartolo, who was in direct contact with us and provided us with the data we needed for our analyses. We would also like to thank Dr. Anders Søgaard for his valuable insights during the revision of this article. 

\bibliography{anthology,custom}
\bibliographystyle{acl_natbib}

\end{document}